\documentclass[letterpaper]{article} 
\usepackage{aaai25}  
\usepackage{times}  
\usepackage{helvet}  
\usepackage{courier}  
\usepackage[hyphens]{url}  
\usepackage{graphicx} 
\usepackage{natbib}  
\usepackage{caption} 
\usepackage{algorithm}
\usepackage{algorithmic}
\usepackage{amsmath}  
\usepackage{subfigure}
\usepackage{tabularx}
\usepackage{booktabs}
\usepackage{multirow}
\usepackage{amsfonts}  
\usepackage{bm}

\usepackage{newfloat}
\usepackage{listings}
\DeclareCaptionStyle{ruled}{labelfont=normalfont,labelsep=colon,strut=off} 
\floatstyle{ruled}
\newfloat{listing}{tb}{lst}{}
\floatname{listing}{Listing}
\title{Path-Adaptive Matting for Efficient Inference under \\Various Computational Cost Constraints}
\author{
    Qinglin Liu \textsuperscript{\rm 1}, Zonglin Li\textsuperscript{\rm 1}, Xiaoqian Lv\textsuperscript{\rm 1}\thanks{Corresponding Author}, Xin Sun\textsuperscript{\rm 1}, Ru Li\textsuperscript{\rm 1}, Shengping Zhang\textsuperscript{\rm 1}\\
    }
\affiliations{
    \textsuperscript{\rm 1} Harbin Institute of Technology, Weihai, China\\

   qinglin.liu@outlook.com, zonglin.li@hit.edu.cn, xiaoqian.hit@gmail.com, \\
   sunxintyc@hit.edu.cn, liru@hit.edu.cn, s.zhang@hit.edu.cn

}

\usepackage{bibentry}

\begin{document}

\maketitle

\begin{abstract}
In this paper, we explore a novel image matting task aimed at achieving efficient inference under various computational cost constraints, specifically FLOP limitations, using a single matting network. Existing matting methods which have not explored scalable architectures or path-learning strategies, fail to tackle this challenge. To overcome these limitations, we introduce Path-Adaptive Matting (PAM), a framework that dynamically adjusts network paths based on image contexts and computational cost constraints. We formulate the training of the computational cost-constrained matting network as a bilevel optimization problem, jointly optimizing the matting network and the path estimator. Building on this formalization, we design a path-adaptive matting architecture by incorporating path selection layers and learnable connect layers to estimate optimal paths and perform efficient inference within a unified network. Furthermore, we propose a performance-aware path-learning strategy to generate path labels online by evaluating a few paths sampled from the prior distribution of optimal paths and network estimations, enabling robust and efficient online path learning. Experiments on five image matting datasets demonstrate that the proposed PAM framework achieves competitive performance across a range of computational cost constraints.
\end{abstract}

\section{Introduction}
Natural image matting is a classic computer vision task, aiming to  estimate the alpha matte of the foreground in a given image. This technique serves as a key technology in many applications such as remote meetings, live streaming, and post-production in film. Hence, it has been extensively researched over the past few decades.
Mathematically, the matting problem aims to estimate the alpha matte \( \alpha \) given an image \( I \) and its foreground \( F \) and background \( B \) as
\begin{equation}
\alpha = \arg \min_\alpha | I - \alpha \cdot F + (1 - \alpha) \cdot B |
\end{equation}
Since only \( I \) is known while \( F \) and \( B \) are both unknown, image matting is an ill-posed problem necessitating additional assumptions or knowledge for resolution.
Traditional matting methods \cite{ wang2007optimized, levin2008a, he2011a} estimate the alpha matte by distinguishing foreground and background or propagating color information, which often struggle in scenarios with overlapping color distributions.
Recently, deep matting methods \cite{qiao2020attention, park2022matteformer,yao2024vitmatte} have employed neural networks to estimate the alpha matte, leveraging high-level semantics to distinguish foreground and background, thereby achieving significant progress.

Despite the remarkable performance achieved by deep image matting methods, real-world image editing applications prioritize achieving efficient inference across different hardware. Unfortunately, most of these methods rely on complex architectures, making them only suitable for GPU deployment. For instance, FBAMatting, a popular matting method~\cite{forte2020fbamatting}, consists of 34.8 million parameters and requires 686 GFlops  to process a $1024 \times 1024$ image, making it impractical for smartphones.
Furthermore, the architectures of current matting methods are non-scalable and require extensive tuning and retraining to achieve optimal performance under different computational cost constraints~\cite{hu2023diffusion,yao2024vitmatte}, leading to substantial computational overhead and carbon dioxide emissions.
While dynamic network methods based on the Gumbel-Softmax  technique~\cite{yu2019any,li2021dynamic_slimmable,HanYPXSSH22,han2024latency} hold promise for constructing scalable matting networks under computational cost constraints, their path-learning strategies largely suffer from the well-known performance collapse issue due to the aggregation of skip connections \cite{xu2019pcdarts,chu2021noisy}.

In this paper, we present  Path-Adaptive Matting (PAM), a framework that learns to dynamically adjust the network path based on image contexts and computational cost constraints, specifically FLOP limitations, for efficient inference. 
Our approach first formulates computational cost-constrained matting network training as a bilevel optimization problem, which optimizes the matting network and the path estimator successively~\cite{anandalingam1992hierarchical,colson2007overview}. 
Based on this formulation, we then design a path-adaptive matting architecture by incorporating path selection layers and learnable connect layers to estimate optimal paths and perform efficient inference within a unified network. 
Then, we propose a performance-aware path learning strategy to generate network path labels online by evaluating only a few network paths sampled from the prior distribution of optimal paths and network estimations, thus enabling robust and efficient online path learning.
Extensive experimental results on five image matting datasets demonstrate that the proposed PAM framework achieves competitive performance across a range of computational cost constraints.

The main contributions of this paper are as follows

\begin{itemize}
\item We present Path-Adaptive Matting (PAM), the first image matting framework that can adaptively adjust the network path based on the image contexts and computational cost constraints for efficient inference.
 
\item We formalize computational cost-constrained matting network training as a bilevel optimization problem, optimizing both the matting network and the path estimator, and design a path-adaptive matting architecture to address it using a unified network.

\item We introduce a performance-aware path learning strategy to generate path labels online by evaluating only a few network paths sampled from the prior distribution of optimal paths and network estimations, thus enabling robust and efficient online path learning.

\item Extensive experiments on five popular image matting datasets demonstrate that the proposed PAM framework achieves competitive performance across a range of computational cost constraints.

\end{itemize}

\section{Related Work}
Researchers have delved into deep image matting methods and efficient neural network design. Here, we provide an overview of the significant related work in this field.

\noindent \textbf{Deep image matting methods.}
Deep image matting methods~\cite{xu2017deep,yu2020mask,dai2021learning,dai2022boosting,liu2024aematter} learn generalizable knowledge from the dataset to estimate alpha matte.
DIM~\cite{xu2017deep} provides the first image matting dataset and introduces an end-to-end matting network, which makes the method robust in a variety of complex scenes.
GCAMatting~\cite{li2020natural} adopts a Guided Contextual Attention module to extract contextual affinity to estimate the alpha matte of semi-transparent objects.
HDMatt~\cite{yu2020high} adopts a cross-patch contextual module to aggregate image contexts for robust patch-based matting.
MGMatting~\cite{yu2020mask} adopts a progressive refinement to estimate the alpha matte from coarse masks, which avoids the laborious trimap annotation.
TIMI~\cite{Liu_2021_ICCV} proposes a 3-branch encoder to mine the global and local neglected coordination, which helps improve predictions on rough trimaps.
LFPNet~\cite{liu2021lfpnet} adopts a center-surround pyramid pooling to propagate contextual information to help process high-resolution images.
RMat~\cite{dai2022boosting} adopts the transformer to aggregate image contexts and explores data augmentation for improving method robustness.
MODNet~\cite{ke2020is} employs a lightweight matting network and a self-supervised strategy to improve the efficiency and robustness.
MatteFormer~\cite{park2022matteformer} integrates CNN and transformer to extract detailed features and long-range features to achieve high performance.
VitMatte~\cite{yao2024vitmatte} employs vision Transformers with a hybrid attention mechanism to improve performance.

\noindent \textbf{Efficient neural network design.}
Early researchers proposed neural architecture search methods \cite{Xie2017Genetic,liu2019darts,chen2019progressive,xu2019pcdarts} to design efficient neural networks under various computational cost constraints. 
However, these methods are often time-consuming due to the need for training and evaluating independent structures. Therefore, recent studies have introduced dynamic neural networks \cite{wang2018skipnet,liu_fastbert_2020,schwartz_right_2020,zhou_bert_2020} that adjust network structures during inference, significantly reducing the cost of network design by training only one supernet.
Early research explored early-exit networks \cite{xin_deebert_2020,schwartz_right_2020,zhou_bert_2020}, which reduce computational costs by leveraging predictions from early layers of the network. MSDNet \cite{huang2017multi} adopts a multi-scale architecture, stopping inference when shallow networks achieve high-confidence results to avoid redundant computation. RANet \cite{yang_resolution_2020} employs intermediate classifiers at different scales within a single network, terminating inference when high-confidence predictions are obtained at lower scales. However, these networks exhibit significant differences in feature representations across scales and are primarily suitable for classification tasks.
Recent research \cite{wang2018skipnet,yu2019any} focuses on enhancing feature continuity through dynamic depth or width adjustment. 
DRNet \cite{caidynamic} employs an undirected graph design to predict the connections of current module nodes, altering network width. 
LASNet~\cite{HanYPXSSH22} introduces a Gumbel-Softmax based mask estimator that guides the network bypass low-value regions, thereby accelerating the inference.
However, these works focus on classification tasks, with limited attention to dense regression at high resolutions, making them less effective for addressing the computational cost constrained matting task explored in this work.

\section{Method}
In this section, we first formalize training a computational cost-constrained matting network as a bilevel optimization problem, where the upper level optimizes the network path estimator and the lower level optimizes the matting network. Based on this formalization, we introduce the Path-Adaptive Matting (PAM) framework to tackle this problem using a unified network. Subsequently,  we delve into a comprehensive introduction of the PAM framework, elucidating its path-adaptive matting architecture and the performance-aware path learning strategy. Finally, we present the loss functions incorporated within the framework.

\begin{figure*}[!t]
    \begin{center}
    \includegraphics[width=0.96\linewidth]{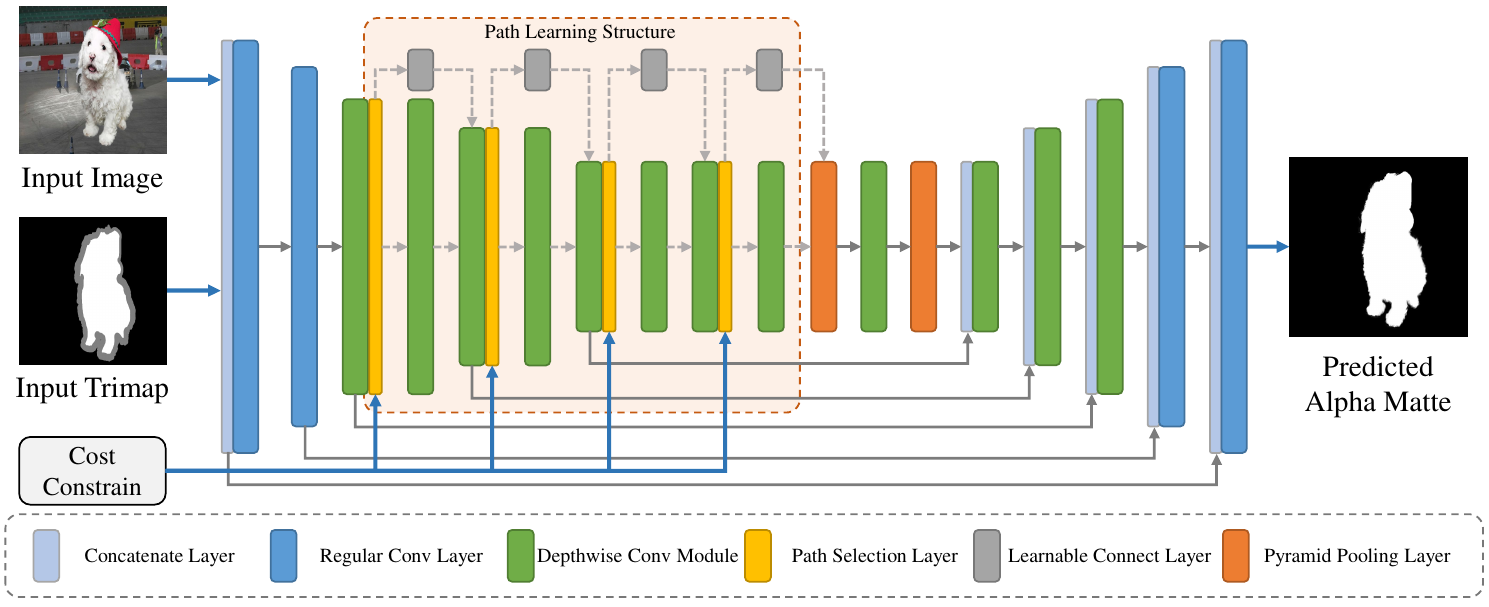}
    \end{center}
    \caption{Overview of the path-adaptive matting architecture. 
We build a lightweight matting backbone using regular and depthwise convolution layers. To enable path-adaptive inference, a path-learning structure is introduced, which uses path selection layers to estimate network paths based on cost constraints and image context, and learnable connect layers for layer bypassing.
    }
    \label{fig:overall}
\end{figure*}

\subsection{Problem Formulation}
The objective of training a computational cost-constrained image matting network is to achieve the minimum matting error while adhering to a specified computational cost constraint $C_c$. This problem can be cast as a bilevel optimization problem: the lower level entails the optimization of weights across all conceivable matting networks, while the upper level optimizes the parameters of an estimator (using a neural network for approximation) to predict the optimal network path. Mathematically, it can be expressed as
\begin{equation}
\begin{aligned}
&\omega_p^*=\arg\min_{\omega_p} \mathbb{E}_{I,C_c} \left[ \mathcal{L} \left( N_m(I,\omega_n^*, N_p(I,C_c,\omega_p)),\alpha^{gt} \right) \right] \\
\quad& \text{s.t.} \quad 
\begin{aligned}
& \omega_n^*=\arg\min_{\omega_n}   \mathbb{E}_{I,P}  \left[ \mathcal{L} \left( N_m(I, \omega_n, P),\alpha^{gt} \right) \right] \\
& \mathrm{F_c} \left( N_p(I,C_c,\omega_p) \right) \leq C_c 
\end{aligned}
\end{aligned}
\label{lb:eq2}
\end{equation}
Here, $\omega_p$ and $\omega_n$ denote the network weights, and $\mathcal{L}(\cdot, \cdot)$ denotes the loss function. $\alpha^{gt}$ represents the ground truth alpha matte. The function $N_m(I, \omega_n, P)$ represents a matting super network that can be inferred with a given path $P$ and network weights $\omega_n$. The function $N_p(C_c, \omega_p)$ represents a path estimation network that can be inferred with a given path $C_c$ and network weights $\omega_p$. The computational cost of the network is denoted by $\mathrm{F_c}(\cdot)$.

Based on Equation~\ref{lb:eq2}, it is evident that the problem involves training two networks. The lower-level task focuses on optimizing the weights $\omega_n$ of the matting network $N_m$ to ensure high matting performance across all paths $P$. Conversely, the upper-level task involves optimizing the weights $\omega_p$ of the path estimation network $N_p$ to predict network paths that adhere to the computational cost constraint $C_c$ while achieving the best performance. However, employing two networks for prediction entails significant computational costs when inferring high-resolution images, thus substantially affecting computational efficiency. Therefore,  we propose that network path estimation can be conducted locally based on the current image context and computational cost constraints. 
Building upon this, we employ a method of predicting the path of the subsequent network stage at each stage of the network, enabling the resolution of the bilevel optimization problem using a unified network. In particular, we design a path-adaptive matting architecture composed of efficient matting network layers and leverage a path learning structure to estimate paths with image contexts and computational cost constraints, thereby dynamically controlling the inference of the matting network.

\subsection{Path-Adaptive Matting Architecture}
As illustrated in Figure~\ref{fig:overall}, the path-adaptive matting architecture first constructs a lightweight matting network. Regular convolution layers are adopted at the lower levels, while the higher levels are based on depthwise convolution modules that consist of large kernel depthwise convolution and point wise convolution, which ensures high computational efficiency and performance.  In addition, we introduce stacked pyramid pooling modules~\cite{zhao2017pyramid} to help aggregate image semantics. Based on the lightweight network, we incorporate cost constraint embedding to input the cost constraint.  We then adopt a path learning structure, which includes path selection layers and learnable connect layers, to estimate optimal paths and perform  efficient inference along the estimated paths.

\subsubsection{Cost Constraint Embedding.} 
To feed computational cost constraint information to the path learning structures, we embed the input computational cost constraint into fixed-length features.
To match the constraints with the network architecture, we first evaluate the smallest and largest sub-networks and obtain their computational costs $C_{min}$ and $C_{max}$, respectively.
Then, we use an embedding layer to convert the integer computational cost constraint $C_c$ between $C_{min}$ and $C_{max}$ into the computational cost constraint feature $\boldsymbol{F}^c$, which is subsequently fed to the network.

\subsubsection{Path Learning Structure.}
The path-selective structure uses path selection layers and learnable connect layersto estimate the optimal path based on given computational cost constraints and image contexts, and to perform learnable layer bypassing, respectively.  Below, we will provide a detailed explanation of these layers.

\textit{ {Path Selection Layer.}} 
To estimate the optimal network path under the given computational cost constraint and image contexts, we design a path selection layer to estimate the classification of the path, i.e., whether to bypass adjacent layers.
Considering that the global semantics can aid in determining the optimal network path and be used to generate attention weights to refine features, we align the path selection layer with a channel attention design. We obtain global semantics by averaging the image features \(\boldsymbol{F}_i^{s}\) from the \(i\)-th layer, and then generate channel attention to refine the network features and use multi-layer perceptron  to estimate  the distribution of the optimal network paths $\boldsymbol{Q}_i^e$ as 
\begin{equation}
\begin{aligned}
\boldsymbol{F}_i^{gs}&= {\rm{Relu}}( {\rm {Conv}} ({\rm {Avgpool}}( \boldsymbol{F}_i^{s}))) \\
\boldsymbol{F}_i^{rf}&=  \boldsymbol{F}_i^{s} \otimes  {\rm{Sigmoid}}({ \rm {Conv} }
(\boldsymbol{F}_i^{gs})) \\
\boldsymbol{Q}_i^e&= {\rm{MLP}}( {\rm {Concat}} (\boldsymbol{F}_i^{gs}, \boldsymbol{F}^c)) 
\end{aligned}
\label{eq:attention}
\end{equation}
Here, \(\boldsymbol{F}_i^{rf}\) denotes the refined semantic features. \(\rm{Conv}(\cdot)\)  \(\rm{MLP}(\cdot)\) and \(\rm{Concat}(\cdot,\cdot)\) denote the convolution, multi-layer perceptron, and concatenation layers, respectively.

 \textit{ {Learnable Connect Layer.}} To enhance network efficiency, we bypass unnecessary paths. However, naive skip connections often behave differently from depthwise convolution modules, leading to discrepancies between the features of bypassed and non-bypassed paths, which can impact performance. To address this challenge, we introduce learnable connect layers designed to approximate the transformation of features from bypassed layers.
The learnable connect layer uses a residual block consisting of two \(1 \times 1\) pointwise convolution layers to process the source feature \(\boldsymbol{F}^{src}\) and output the destination feature \(\boldsymbol{F}^{des}\) as
\begin{equation}
\begin{aligned}
\boldsymbol{F}^{des} = {\rm{Conv}}({\rm{ReLU}}({\rm{Conv}}(\boldsymbol{F}^{src}))) + \boldsymbol{F}^{src}
\end{aligned}
\end{equation}
Since our learnable connect layer only involves pointwise convolutions, it is more computationally efficient than the bypassed depthwise convolution modules.

\subsection{Performance-Aware Path Learning}
With the path-adaptive matting architecture, our objective is to train the network to estimate the optimal path and perform efficient inference.
 However, we have empirically found that using the Gumbel-Softmax trick~\cite{HanYPXSSH22,han2024latency} often leads the path selection layers to favor the shortest path. 
This is because gradient-based NAS methods inherently suffer from operation co-adaptation issues, which often lead to an excessive aggregation of skip connections.
Conversely, while training the network with the optimal path yields good results, obtaining the ground truth distribution requires evaluating all paths in each iteration, which is computationally expensive.
To tackle these issues, we propose a performance-aware path learning strategy that generates path labels through online evaluation of paths generated by a prior distribution of the optimal path and the network predictions. 
Specifically, we use the Monte Carlo to estimate a prior distribution of the optimal path as a path generator.
During training, we use an online path label generation method as described in Algorithm~\ref{alg:generation}, evaluating both the paths generated from the prior distribution and those predicted by the network. The path with the highest performance is used as the label.

\begin{algorithm}[t]
\caption{Performance-Aware Path Learning}
\label{alg:generation}
\begin{algorithmic}[1]
\STATE \textbf{Input:} Paths $\mathbf{R}= \{\boldsymbol{R}_1, \boldsymbol{R}_2, \dots, \boldsymbol{R}_{N^e}\}$ generated by the prior distribution and the matting errors $\mathbf{E}=\{E_1, E_2, \dots, E_{N^e}\}$, path $\boldsymbol{V}$ estimated by the network and the corresponding matting error $E^v$ and computational cost $C^v$, the given computational cost constraint $C^g$.
\STATE \textbf{Output:} Optimal path $\boldsymbol{R}^o$.
\STATE Initialize the optimal path $\boldsymbol{R}^o \gets \emptyset$, the lowest matting error $E^o \gets +\infty$
\IF {$C^v < C^g$}
    \STATE $\boldsymbol{R}^o \gets \boldsymbol{V}$
    \STATE $E^o \gets E^v$
\ENDIF
\FOR {$i \gets 1$ to $N^e$}
    \IF {$E_i < E^o$}
        \STATE $\boldsymbol{R}^o \gets \boldsymbol{R}_i$
        \STATE $E^o \gets E_i$
    \ENDIF
\ENDFOR
\end{algorithmic}
\end{algorithm}

\subsubsection{Prior Distribution Estimation.}
Given a computational cost constraint $C_c$ and an input image $\boldsymbol{I}$,  the ground truth distribution $\boldsymbol{Q}^{gt} = {\rm P}(X_1,X_2\dots,X_{N^a}|\boldsymbol{I},\boldsymbol{F}^c)$ of optimal paths can only be obtained by evaluating all paths in each iteration, which is computationally unaffordable.
To solve this problem, we use a distribution that approximates $\boldsymbol{Q}^{gt}$ as a prior distribution to generate a few optimal candidate paths for efficient path evaluation.
We compare a uniform distribution $\boldsymbol{Q}^{u}$ and a distribution $\boldsymbol{Q}^{c}  = {\rm P}(X_1,X_2\dots,X_{N^p}| \boldsymbol{F}^c)$ of the optimal path under the given computational cost constraint.
The expectation of ${p}(X_1^r,X_2^r\dots,X_{N^a}^r|\boldsymbol{I},\boldsymbol{F}^c)$ is 
\begin{equation}
\mathbb{E}({p}(X_1^r,X_2^r\dots,X_{N^a}^r|\boldsymbol{I},\boldsymbol{F}^c))=\frac{1}{N^{ap}}
\end{equation}
where $X_1^r,X_2^r\dots,X_{N^a}^r$  is the path   sampled from the uniform distribution $\boldsymbol{Q}^{u}$, $N^{ap}$ is the number of all possible paths.
The expectation of ${p}(X_1^c,X_2^c\dots,X_{N^a}^c|\boldsymbol{I},\boldsymbol{F}^c)$ is 
\begin{equation}
\mathbb{E}({p}(X_1^c,X_2^c\dots,X_{N^a}^c|\boldsymbol{I},\boldsymbol{F}^c))={p}(X_1^o,X_2^o\dots,X_{N^a}^o| \boldsymbol{F}^c)
\end{equation}
where $X_1^c,X_2^c\dots,X_{N^a}^c$ is the path sampled from the distribution $\boldsymbol{Q}^{c}$ and
$X_1^o,X_2^o\dots,X_{N^a}^o$ is the optimal path. 
Since ${p}(X_1^o,X_2^o\dots,X_{N^a}^o|\boldsymbol{F}^c)$ is usually much larger than $\frac{1}{N^{ap}}$, the path sampled from the distribution $\boldsymbol{Q}^{c}$ is more likely to be optimal.
Therefore, we adopt the distribution $\boldsymbol{Q}^{c}$ as the prior distribution and use the Monte Carlo method to estimate the distribution via simulation.

To obtain the distribution $\boldsymbol{Q}^{c}$, we first follow SPOS~\cite{guo2019single} to train matting networks of random paths in the path-adaptive matting architecture.
Then, we define a probability estimator that the path $X_1,X_2\dots,X_{N^a}$ is the optimal path under the computational cost constraint $C_{g}$ as
\begin{equation}
\widehat{p}^g_{N^{val}}(X_1,X_2\dots,X_{N^a})=\frac{1}{N^{val}}\sum_{i=1}^{N^{val}}{{B_i}}
\end{equation}%
where $N^{val}$ denotes the number of input images. $B_i\in\{0, 1\}$ denotes whether the path $X_1,X_2\dots,X_{N^a}$ is optimal or not for the $i$-th image and the computational cost constraint  $C_{g}$.
Next, we simulate $N^{val}$ images using the foreground and background images in the dataset.
Since the computational cost is positively correlated with the network performance, we simulate $N^g$ paths whose computational costs are close to the given computational cost constraint $C_g$ to reduce computational costs.
Finally, we use a pretrained whole matting network to evaluate the simulated images and paths and obtain the estimate $\widehat{\boldsymbol{Q}}^{c}$ of the distribution as the prior distribution.

\begin{figure*}[!t]
\centering
		\includegraphics[width=0.99\textwidth]{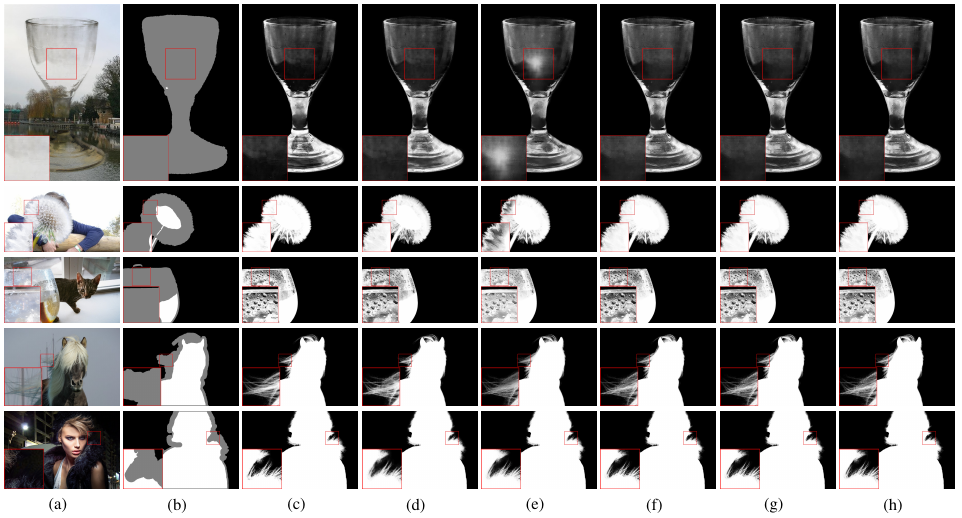}
    \caption{Qualitative results on the Adobe Composition-1K dataset. (a) Input Image. (b) Trimap. (c) Ground Truth. (d) GCAMatting. (e) MatteFormer. (f)  PAM (Aggressive). (g) PAM  (Moderate). (h) PAM  (Mild). }
    \label{fig:adb}
\end{figure*}

\begin{table*}[!t]
  \centering
    \begin{tabularx}{\linewidth}{p{5cm}|>{\centering} >{\centering\arraybackslash}X>{\centering\arraybackslash}X>{\centering\arraybackslash}X>{\centering\arraybackslash}X|>{\centering\arraybackslash}X>{\centering\arraybackslash}X>{\centering\arraybackslash}X>{\centering\arraybackslash}X}
    \toprule
    Method&  {SAD} & {MSE} & {GRAD} & {CONN} & {Flops} & {Param}& {Latency}& {Memory} \\
    \midrule
    DIM~\cite{xu2017deep}  &50.40  & 17.00  & 36.70  & 55.30  & 727.4G   & 130.5M& 11.97s & 18 GB  \\
    IndexNet~\cite{lu2019indices}& 45.80  & 13.00  & 25.90  & 43.70  & 116.6G     & 8.2M & 7.67s  & 8 GB \\
    GCAMatting~\cite{li2020natural}  & 35.28  & 9.00  & 16.90  & 32.50  & 257.3G   & 24.1M  & 37.76s & 13 GB\\
    FBAMatting~\cite{forte2020fbamatting} & 26.40  & 5.40  & 10.60  & 21.50  & 686.0G    & 34.8M& 15.44s & 13 GB   \\
    SIM~\cite{sun2021sim}  & 27.70  & 5.60  & 10.70  & 24.40  & 1001.9G  & 44.5M  & 20.17s & 14 GB \\
    TIMI~\cite{Liu_2021_ICCV} & 29.08  & 6.00  & 11.50  & 25.36  & 351.3G    & 33.5M & 124.07s & 72 GB \\
    LFPNet~\cite{liu2021lfpnet} & 23.60  & 4.10  & 8.40  & 18.50  & 1539.4G    & 112.2M & 45.93s & 16 GB \\
    MatteFormer~\cite{park2022matteformer}  & 23.80  & 4.00  & 8.70  & 18.90  & 233.3G    & 44.9M & 14.39s & 11 GB \\
            DiffusionMat~\cite{xu2023diffusionmat} & 22.80	&4.00&	6.80&	18.40&29212.5G&32.8M&551.00s&33GB\\
                VitMatte~\cite{yao2024vitmatte} & 20.33	&3.00&	6.74&	14.78&784.8G&89.2M&130.71s&30GB \\
    \midrule
    PAM (Mild) &   23.14    &  4.17&   8.48    &     18.57  & 74.6G & 7.1M& 6.25s  & 8 GB \\
    PAM (Moderate) &   23.25   &   4.25    &  8.50     &18.71       & 69.6G  & 7.1M& 5.58s  & 8 GB\\
    PAM (Aggressive)&   24.05     &    4.52   &  8.97     & 19.63       &  57.8G   & 7.1M & 5.23s  & 8 GB\\
    \bottomrule
    \end{tabularx}%
                    \caption{Quantitative results on Adobe Composition-1K. Our PAM method is evaluated under three computational cost constraints: mild (under 75 GFlops), moderate (under 70 GFlops), and aggressive (under 60 GFlops). Flops denotes the floating-point computations required for inferring a $1024 \times 1024$ image. Param denotes the network parameter number. Latency and memory refer to the latency and peak memory usage measured on an R9 3900X CPU for inferring a $2048 \times 2048$ image.}
  \label{tab:adbq}%
\end{table*}%

\subsubsection{Online Path Label Generation.}
Although the prior distribution provides good path candidates, using it as a label limits the network's ability to search for better paths. 
Inspired by heuristic algorithms, we propose an online path label generation method, where we assess the paths generated by the prior distribution and those predicted by the network. The best-performing path among these is then used as the label for training.
In each iteration of the training phase, we first generate candidate paths for the given computational constraint $C^g$ with the prior distribution.
Then, we use the distribution $\widehat{\boldsymbol{Q}}^{c}$ to randomly sample $N^e$ candidate paths ${\mathbf{R}}= \{\boldsymbol{R}_1, \boldsymbol{R}_2, \dots, \boldsymbol{R}_{N^e}\}$ without repetition, and evaluate the sampled paths with the L1 loss function to obtain the errors  $\mathbf{{E}}=\{{E}_1, {E}_2, \dots,  {E}_{N^e}\}$.
Next, we use the network itself to sample the path $\boldsymbol{V}$  by independently estimating the path $\boldsymbol{V}_i$  for the $i$-th path selection layer with the unnormalized distribution $\boldsymbol{Q}_i^e$ as 
\begin{equation}
{V}_{i,j}=\frac{\exp(({{Q}^e_{i,j}}+{G}_{i,j})/\tau)}{\sum_{k=1}^2 \exp(({{Q}^e_{i,k}}+{G}_{i,k})/\tau)}
\end{equation}
where $\boldsymbol{G} \in \mathbb{R}^{N^a \times 2}$ is a Gumbel noise that helps the network to explore low probability paths.
${V}_{i,j}$, ${Q}^e_{i,j}$, and ${G}_{i,j}$ are the $j$-th items of $\boldsymbol{V}_i$, $\boldsymbol{Q}_i^e$, and $\boldsymbol{G}_i$, respectively.
$\tau$ is the temperature coefficient that controls the smoothness of the result.
In addition, we use a gumbel-max reparameterization trick~\cite{huijben2022review} to convert the estimated path $\boldsymbol{V}_i$ to a one-hot tensor while keeping it differentiable.
We evaluate the matting network with the path $\boldsymbol{V}$ to obtain the error ${E}^v$ and the computational cost $C^v$.
Finally, we use the evaluation algorithm described in Algorithm~\ref{alg:generation} to identify the path with the lowest error as the online path label $\boldsymbol{R}^o$, which is subsequently  used to train the network.

\subsection{Loss Functions}
To train the PAM framework, we define the network loss as
\begin{equation}
\begin{aligned}
\mathcal{L} = \lambda_{\alpha}\mathcal{L}_{\alpha} + \lambda_{ds}\mathcal{L}_{ds}+ \lambda_{pt}\mathcal{L}_{pt}
\end{aligned}
\end{equation}
where $\mathcal{L}_{\alpha}$ is the alpha matte loss defined as
\begin{equation}
\label{eq:alpha}
\begin{aligned}
\mathcal{L}_{\alpha} = \lambda_{1}\mathcal{L}_{1} + \lambda_{comp}\mathcal{L}_{comp}+ \lambda_{lap}\mathcal{L}_{lap}
\end{aligned}
\end{equation}
where $\mathcal{L}_{1}$, $\mathcal{L}_{comp}$, and $\mathcal{L}_{lap}$  are the L1 loss,  compositional loss,  and Laplacian loss, as defined in MatteFormer~\cite{park2022matteformer}.  
$\lambda_{1}$, $\lambda_{comp}$, and  $\lambda_{lap}$ are the weights for the three losses. 
The distillation loss $\mathcal{L}_{ds}$ is designed to help the learnable connect layers to learn the feature transformation of the bypassed layer with the labels generated by a trained complete PAM network, which is defined as 
\begin{equation}
\mathcal{L}_{ds} = \frac{1}{|\mathcal{T}^U|}\sum_{{i\in\mathcal{T}^U}}{\sqrt{(\alpha_i-\alpha_i^{sd})^2+\epsilon^2}}
\end{equation}
where $\alpha_i^{sd}$ is the alpha matte of the $i$-th pixel estimated by a trained complete PAM network. $\mathcal{T}^U$ is a set of unknown pixels in the trimap. 
$\epsilon$ is the penalty coefficient. 
The path loss $\mathcal{L}_{pt}$ is designed to use the optimal path $\boldsymbol{R}^o$ to supervise the predictions of path selection layers $\boldsymbol{Q}^e$ as
\begin{equation}
\mathcal{L}_{pt} =\sum_{i=1}^{N^a} { \rm {CrossEntroy}} (\boldsymbol{Q}_i^e,\boldsymbol{R}_i^o)
\end{equation}
where $\rm {CrossEntropy}(\cdot,\cdot)$ denotes cross entropy function.

\section{Experiments}
In this section, we first present the implementation details of the proposed PAM framework. Next, we evaluate PAM against existing methods using synthetic datasets, including Adobe Composition-1k~\cite{xu2017deep}, Distinctions-646~\cite{qiao2020attention}, Transparent-460~\cite{cai2022TransMatting}, Semantic Image Matting (SIMD)~\cite{sun2021sim}, as well as the real-world Automatic Image Matting-500 (AIM-500) dataset~\cite{ijcai2021li}.

\subsection{Implementation details}
The proposed method is implemented using the PyTorch framework. We train our PAM framework on an RTX 3090 GPU with a batch size of 4. 
All network weights are initialized using the Kaiming initializer~\cite{he2015delving}. 
To avoid overfitting, we follow the data preprocessing methods of previous matting methods to process the train data~\cite{forte2020fbamatting}.
The training process is divided into three stages. 
The networks are trained using the Radam optimizer~\cite{liu2019radam} with a weight decay of $3 \times 10^{-5}$ and betas of $(0.5, 0.999)$. The initial learning rate is set to $3 \times 10^{-4}$ and decays to zero using a cosine annealing scheduler in each stage.
In the first stage, we train the entire PAM network for 150 epochs. In the second stage, we perform warm-up training by randomly sampling sub-networks and training them for 20 epochs. In the third stage, we train PAM with the performance-aware path learning strategy for 150 epochs.
The other coefficients used in this method are configured as follows:  $N^a = 4$,  $\lambda_{\alpha} = 1$, $\lambda_{ds} = 0.05$, $\lambda_{pt} = 0.05$, $\lambda_1 = 1$, $\lambda_{comp} = 0.25$, $\lambda_{lap} = 0.5$, $\epsilon=10^{-6}$, $N^e = 4$, $N^{val} = 10^3$, $N^g = 10$, and $\tau = 1$.

\begin{table*}[]
	\centering
    \begin{tabularx}{\linewidth}{p{5cm}|>{\centering} >{\centering\arraybackslash}X>{\centering\arraybackslash}X>{\centering\arraybackslash}X>{\centering\arraybackslash}X|>{\centering\arraybackslash}X>{\centering\arraybackslash}X>{\centering\arraybackslash}X>{\centering\arraybackslash}X}
		\toprule
		\multirow{2}{*}{Method} & \multicolumn{4}{c}{Distinctions-646}             & \multicolumn{4}{|c}{Transparent-460}             \\  \cmidrule(lr){2-9}     
		                        & SAD        & MSE        & GRAD       & CONN       & SAD        & MSE        & GRAD       & CONN       \\ \midrule
		DIM~\cite{xu2017deep}                     &      56.13 & 22.84 & 50.01 & 57.90      &    356.20& 49.68& 146.46 &296.31         \\
		IndexNet~\cite{lu2019indices}                &    40.31 &8.23 & 37.60 &39.92    &   434.14 &74.73 &124.98& 368.48      \\
		GCAMatting~\cite{li2020natural}              &     31.32 &6.64 & 28.69 &30.45       &    219.38        &      23.17      &     130.46       &   224.65         \\
  TIMI~\cite{Liu_2021_ICCV} & 42.61 & 7.75& 45.05 & 42.40 & 328.08& 44.20 &142.11& 289.79\\
		MGMatting~\cite{yu2020mask}                        &  33.24&  4.51&  20.31 & 25.49      &    344.65& 57.25 &74.54& 282.79   \\
  TransMatting~\cite{cai2022TransMatting} &25.65 &3.40 &16.08&21.45&    192.36& 20.96& 41.80 &158.37\\
   \midrule
        PAM (Mild) & 21.97 & 3.45 & 15.80 & 20.65 &194.09& 20.27&34.26& 197.35    \\
            PAM (Moderate) & 22.02 & 3.44 & 15.73 & 20.72&199.24&21.60&34.30&203.65 \\
            PAM (Aggressive) &22.97 & 3.78& 16.31 & 21.90  &210.11&23.55&41.42&215.95\\  
            \bottomrule
	\end{tabularx}
 	\caption{Quantitative results on Distinctions-646 and Transparent-460.}
	\label{tab:646460}
\end{table*}

\begin{table*}[]
	\centering
    \begin{tabularx}{\linewidth}{p{5cm}|>{\centering} >{\centering\arraybackslash}X>{\centering\arraybackslash}X>{\centering\arraybackslash}X>{\centering\arraybackslash}X|>{\centering\arraybackslash}X>{\centering\arraybackslash}X>{\centering\arraybackslash}X>{\centering\arraybackslash}X}
		\toprule
		\multirow{2}{*}{Method} & \multicolumn{4}{c}{Semantic Image Matting}             & \multicolumn{4}{|c}{Automatic Image Matting-500}             \\  \cmidrule(lr){2-9}  
		                        & SAD        & MSE        & GRAD       & CONN       & SAD        & MSE        & GRAD       & CONN       \\ \midrule
		DIM~\cite{xu2017deep}                     &      98.95&61.34&32.19&103.86     &   47.81 &79.47&38.16&49.07       \\
		IndexNet~\cite{lu2019indices}                &    62.90&25.00&21.76&63.43   &  26.85&26.22&16.37&26.15  \\
		GCAMatting~\cite{li2020natural}              &    72.18&29.82&23.88&71.52     &        34.81&38.93&25.72&35.14         \\
		 LFPNet~\cite{liu2021lfpnet}                        &   22.11&4.32&6.82&17.06   &      26.15&21.14&14.93&25.73          \\
    MatteFormer~\cite{park2022matteformer} & 23.59&4.88&7.67&18.69 & 26.87&29.00&23.00&26.63 \\
   \midrule
    PAM (Mild) & 24.39&4.73&7.52&20.24 &25.14 & 23.73 & 19.74 &25.02    \\
    PAM (Moderate) & 24.68&4.85&7.56&20.54&  25.12 & 23.76 & 19.79 & 25.08 \\
    PAM (Aggressive) &25.37&5.25&7.95&21.42  &24.35&23.47&19.82&24.33\\  
    \bottomrule
	\end{tabularx}
  	\caption{Quantitative results on Semantic Image Matting and Automatic Image Matting-500.}   
	\label{tab:aimq}
\end{table*}

\subsection{Results on the Synthetic Datasets}
To evaluate the performance of our PAM framework, we compare PAM with state-of-the-art methods, including  DIM~\cite{xu2017deep}, IndexNet~\cite{lu2019indices}, GCAMatting~\cite{li2020natural}, FBAMatting~\cite{forte2020fbamatting}, SIM~\cite{sun2021sim}, TIMI~\cite{Liu_2021_ICCV}, LFPNet~\cite{liu2021lfpnet}, MatteFormer~\cite{park2022matteformer}, DiffusionMat~\cite{xu2023diffusionmat}, and VitMatte~\cite{yao2024vitmatte} on the Adobe Composition-1K dataset.
Table~\ref{tab:adbq} provides a summary of the quantitative results for all compared matting methods.
Our PAM is evaluated under three computational cost constraints: mild (under 75 GFlops), moderate (under 70 GFlops), and aggressive (under 60 GFlops). 
Flops denote the floating-point computations required for inferring a $1024 \times 1024$ image. Param represents the number of network parameters. 
Latency and memory refer to the inference latency and memory usage, respectively, measured on an AMD R9 3900X CPU for inferring a $2048 \times 2048$ image.
Figure~\ref{fig:adb} illustrates the qualitative results of PAM compared with other methods.
In comparison with existing methods, our demonstrates competitive performance with significantly lower computational costs and fewer parameters.
PAM under the moderate constraint performs comparably to state-of-the-art image matting methods including LFPNet and MatteFormer.
However, PAM under the moderate constraint consumes only $4.5\%$ computation and $6.3\%$ parameters of LFPNet, $29.8\%$   computation and $15.8\%$ parameters of MatteFormer.

To evaluate the generalization ability of PAM, we compare it with existing matting methods such as IndexNet, GCAMatting, TIMI, TransMatting~\cite{cai2022TransMatting}, and MatteFormer on three synthetic datasets including Distinctions-646, Transparent-460, and Semantic Image Matting.
All compared methods are trained on the Adobe Composition-1K dataset.
Quantitative results are summarized in Tables~\ref{tab:646460} and~\ref{tab:aimq}.
DIM, IndexNet, GCAMatting, LFPNet, MatteFormer, and our PAM exhibit strong performance across all three datasets. Notably, PAM achieves performance on par with leading methods like TransMatting and MatteFormer, indicating its robust generalization ability. Furthermore, as computational cost constraints relax, the performance of PAM also improves, underscoring the effectiveness of our proposed method.

\subsection{Results on the Real-world Dataset}
To evaluate the performance of the proposed PAM framework on real-world images, we use the Automatic Image Matting-500 (AIM-500) dataset to evaluate DIM, IndexNet, GCAMatting,  LFPNet, MatteFormer, and our PAM, which are trained on the Adobe Composition-1K dataset.
Table~\ref{tab:aimq} summarizes the quantitative results of all compared methods.
Compared with existing methods, PAM demonstrates strong generalization ability when applied to real-world images. 
PAM under the mild computational cost constraint has a similar performance to LFPNet and MatteFormer, which suggests that PAM has good generalization ability on real-world images.
However, PAM under the aggressive constraint outperforms PAM under the mild constraint, which may be due to the domain shift when testing PAM on real data, as it is trained on synthetic data.

\section{Conclusion}
In this paper, we propose Path-Adaptive Matting (PAM), a framework that dynamically adjusts network paths based on  image contexts and computational cost constraints for efficient inference. We first formulate the training of the computational cost-constrained matting network as a bilevel optimization
problem, optimizing both the matting network and the path estimator. We introduce a path-adaptive matting architecture by incorporating path selection layers and learnable connect layers to estimate optimal paths and perform efficient inference within a unified network. Furthermore, we propose a performance-aware path learning strategy to generate path labels by evaluating a few paths sampled from the prior distribution of optimal paths and network estimations, thus enabling robust and efficient online path learning.
Experimental results on popular image matting datasets demonstrate that PAM achieves competitive performance across various computational cost constraints.

\bibliography{aaai25}

\end{document}